\documentclass[conference]{IEEEtran}
\IEEEoverridecommandlockouts

\usepackage{graphicx}
\usepackage{amsmath}
\usepackage{algorithm,algorithmic,color}
\usepackage{cite}
\usepackage{enumerate}
\usepackage[shortlabels]{enumitem}

\author{\IEEEauthorblockN{Yi Shi and Yalin E. Sagduyu}  
}

\title{Jamming Attacks on Federated Learning in Wireless Networks 
\thanks{This effort is supported by the U.S. Army Research Office under contract W911NF-20-C-0055. The content of the information does not necessarily reflect the position or the policy of the U.S. Government, and no official endorsement should be inferred.}}
	
\begin{document}

\newcommand{\argmax}{\arg\!\max}

\maketitle

\begin{abstract}
Federated learning (FL) offers a decentralized learning environment so that a group of clients can collaborate to train a global model at the server, while keeping their training data confidential. This paper studies how to launch over-the-air jamming attacks to disrupt the FL process when it is executed over a wireless network. As a wireless example, FL is applied to learn how to classify wireless signals collected by clients (spectrum sensors) at different locations (such as in cooperative sensing). An adversary can jam the transmissions for the local model updates from clients to the server (uplink attack), or the transmissions for the global model updates the server to clients (downlink attack), or both. Given a budget imposed on the number of clients that can be attacked per FL round, clients for the (uplink/downlink) attack are selected according to their local model accuracies that would be expected without an attack or ranked via spectrum observations. This novel attack is extended to general settings by accounting different processing speeds and attack success probabilities for clients. Compared to benchmark attack schemes, this attack approach degrades the FL performance significantly, thereby revealing new vulnerabilities of FL to jamming attacks in wireless networks.
\end{abstract}


\section{Introduction}

In federated learning (FL), multiple clients collaborate to train a machine learning (ML) model under the orchestration of a server. A single client may not have sufficient processing power and sufficient training data to train a good ML model with high accuracy. FL aims to remove these limitations by utilizing multiple devices (each, utilizing its own training data in a confidential manner) in a client-server model \cite{McMahan17:FL}. Clients train their local models using local (private) data and the server combines local models (e.g., by federated averaging (FedAvg)) to obtain a global model. Once the global model is updated, the server sends this model to clients so that they can update their local models. This process can be run several rounds until a termination condition (e.g., the number of rounds or the convergence of global model accuracy) is met. FL has a number of advantages, including low processing requirement per client, high reliability (global model achieves high accuracy even if local models of clients without FL may have low accuracy), privacy (clients send only their local models and do not share their local data), and communication efficiency (the amount of data for models is much smaller than local data) \cite{Survey0:FL, Survey1:FL, Survey2:FL, Survey3:FL}. Given these advantages, FL has found applications in different domains including wireless systems such as mobile edge networks \cite{MEC:FL}, Internet of Things (IoT) \cite{FL:IoT}, 5G \cite{5G:FL}, and 6G \cite{6G:FL}, and can be executed over wireless networks \cite{Wireless5:FL, Wireless4:FL, Wireless1:FL, Wireless2:FL, Wireless3:FL} while accounting for wireless factors such as the possibility of packet errors and the availability of wireless resources

In general, ML is known to be susceptible to manipulations of the inputs in training and test times, as commonly studied under adversarial machine learning (AML) \cite{AMLGeneral}. AML attacks have been applied to the wireless domain \cite{SurveyAML1, SurveyAML2, SurveyAML3}. These attacks include inference (exploratory) attacks \cite{Shi2018, terpek, NetSliceAttack}, evasion (adversarial) attacks \cite{Larsson, Larsson2, Yalin2019, Headley2019, Headley2019-1, Headley2020, Silvija2019, Silvija2019-2, Silvija2019-3, Kim, Kim2, Mao2021, Kim4, Kim5, Larsson3, Kim6, sahay2021deep, Bahramali, Jinho, KimRIS, PowerControl}, poisoning (causative) attacks \cite{Sagduyu1, YiMilcom2018, Luo2019, Luo2020, Luo2021}, Trojan attacks \cite{Davaslioglu19}, spoofing attacks \cite{Shi2019, Shi2021, 5Gbc, IoTGAN}, membership inference attacks \cite{MIA1,MIA2}, and attacks to facilitate covert communications \cite{Gunduz1, Gunduz2, Kim3}. FL itself is vulnerable to various insider exploits such as data poisoning (a malicious client may manipulate its training data, including both labels or features), model update poisoning (a malicious client may send manipulated models to the server), free-riding attack (a malicious client may claim little to none training data to contribute to FL while receiving the global model without contributing much from its local model), and inference of class representatives, memberships, and training inputs and labels \cite{MEC:FL, Mittal, Lyu20:FL-attack, FLSecurity}.

In this paper, we study jamming as an external threat to FL when FL is executed over a wireless network (where a server and multiple clients communicate over wireless channels). As a wireless example, FL is used to train a signal classifier based on the I\&Q data collected by different clients (namely, spectrum sensors) at different locations (corresponding to a cooperative sensing setting). Suppose that there are background transmissions with different waveforms (e.g., BPSK or QPSK) such as in the case of primary and secondary users operating in a dynamic spectrum sharing environment. One emerging example for spectrum coexistence is in the Citizens Broadband Radio Service (CBRS) band \cite{FCCRule} at 3.5GHz that is originally dedicated to incumbent (primary) users such as radar and now is opened to the use by 5G. In this setting, the environmental sensing capability (ESC) network (with distributed sensors) needs to detect (potentially with the use of ML) when the incumbent operation is active or idle (namely, distinguish the radar signals from other signals) \cite{CBRS}.

The received signals are affected by the channel gain and phase shift of the channel between the background transmitter and each sensor (namely, each client in the context of FL). The signal classifier should be general (not specific to a particular sensor) so that it can work for signals received by sensors at different locations. We use multiple sensors as clients to collect their local data samples individually and train their local models. There is also a server that trains the global model as a general signal classifier. When FL is applied in wireless network setting, each client and the server correspond to an individual node with transmit and receive capabilities. We show that this decentralized learning scheme in the context of FL can build the global model for the classifier that achieves high accuracy in test time.

\begin{figure}
  \centering
  \includegraphics[width=0.8\columnwidth]{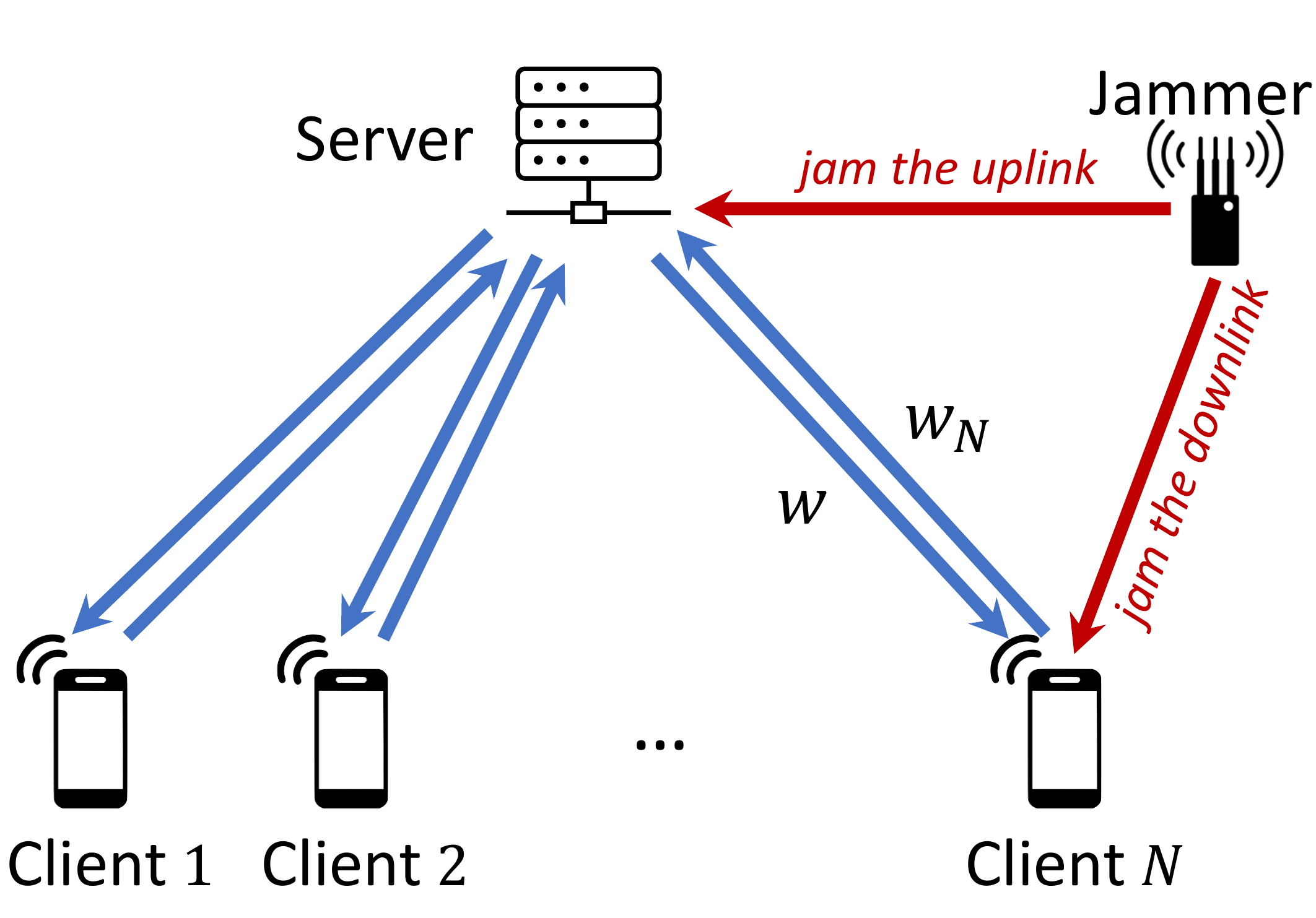}
  \caption{The FL process and the jamming attack on FL in a wireless network setting.}
  \label{fig:FL}
\end{figure}

While vulnerabilities of FL have been extensively discussed in terms of model or data inference and manipulation  \cite{Lyu20:FL-attack, FLSecurity}, the application of FL in the wireless network environment raises a new attack surface.
In particular, the FL process can be attacked by jamming, where an adversary can jam either the uplink updates (local models sent from clients to the server) or the downlink updates (global model sent from the server to clients), or both. The uplink attack prevents some client models from reaching the server such that the global model may not be trained to be general for different clients (i.e., sensors at different locations in the signal classification case). Under the downlink attack, some clients do not receive updates from the server and thus their local models may not converge, which in turn prevents the global model from converging. The adversary may also perform both uplink and downlink attacks (at the same time). We consider an attack budget for the adversary, which can be measured by the average number of attacked clients per round. This budget may be imposed due to the energy constraint of the adversary (namely, for energy efficiency) or its need to avoid jammer detection (namely, for stealthiness). The adversary aims to reduce the accuracy of the trained global model without exceeding the attack budget. For that purpose, the adversary needs to select which set of clients to attack (i.e., which transmissions for model updates to jam). The FL process and the jamming attack on FL in a wireless network setting is illustrated in Fig.~\ref{fig:FL}.

To illustrate the contributions of this paper, we summarize the results obtained based on the analysis of this problem formulation:
\begin{enumerate}
\item We can sort the clients based on the global model accuracy (from worst to best) when a client is under uplink or downlink attack. We can also sort clients based on their local model accuracies (from worst to best) when there is no attack. These two lists are almost the same if the attack is an uplink attack while these two lists are almost in the reverse order if the attack is a downlink attack. The difference between selected clients by these orders is bounded by a small number (e.g., $2$ for uplink attack and $3$ for downlink attack for $10$ clients).

\item The first set of results above is for the accuracy at the end of the FL process. We further find that the list based on local model accuracies converges very fast over time. Thus, this list can be determined after only several (e.g., $3$) rounds of the FL process. The first two sets of results can be used to develop an attack scheme to select clients for uplink, downlink, or simultaneous uplink and downlink attacks under a given budget by comparing the local and global accuracies after several rounds of the FL process. We show that this attack scheme can achieve better performance than the attack with randomly selected clients. For example, the downlink attack on one client by the designed scheme reduces the model accuracy to $63.1$\% while the model accuracy under the random downlink attack remains high ($95.1$\%).

\item There is no limitation that the same number of clients should be attacked in every FL round. We design an attack scheme that can select up to $K$ clients to jam in a round and keep the average number of selected clients as the budget $M$. We observe that that this $(M,K)$ scheme falls behind the $(M,M)$ scheme (i.e., fixed set of $M$ clients attacked for every round) in terms of the reduced global model accuracy under attack (e.g., $93.6$\% for the $(M,M)$ scheme vs. $63.1$\% for the $(M,K)$ scheme for the case of $10$ clients).
\end{enumerate}

We also extend the jamming attack on FL to more general problem settings:
\begin{enumerate}
	\item  Clients may have different processing powers and slow clients may not be able to participate the FL process (by sending local model updates) in every round. In this setting, an adversary does not need to attack all selected clients in every round. Instead, it is necessary to attack a client only if it has a local update and will receive a global update. Thus, the consumed budget for a client can be less than one transmission per round. We extend the attack scheme to correctly account for the consumed budget with consideration of different processing powers. For performance evaluation, we consider four benchmark attack schemes: (i) The adversary jams all selected clients in every round.
	(ii) The adversary first selects fast clients and then slow clients for the attack. (iii) The adversary first selects slow clients and then fast clients for the attack. (iv) The adversary randomly selects clients for the attack. We show that the designed scheme is better than all benchmark schemes. For example, the uplink attack on four clients  selected by the designed scheme reduces the model accuracy to $72.3$\% while the model accuracy under the benchmark attack schemes remains as high as $93.4$\%.
	
	\item The jamming attack may not be always successful, i.e., it is possible that the jamming attack may not always make client or server transmissions fail due to channel effects. We model this case by considering a success probability associated with uplink and downlink attacks depending on channel effects. In this setting, we need to consider both model accuracy and attack success probability to select which clients to attack. For the uplink attack, we identify a metric based on diversity of remaining client local models to select clients and analyze this metric as a function of model accuracy and success probability. For the downlink attack, we use the normalized accuracy (accuracy divided by success probability) to select clients. For performance evaluation, we consider three benchmark attack schemes: (i) The adversary selects clients based on model accuracy only. (ii) The adversary selects clients based on attack success probability only (clients with high probability are selected). (iii) The adversary randomly selects clients to attack. We show that the designed attack scheme is better than all benchmark schemes. For example, attacking both uplink and downlink on seven clients by the designed scheme reduces the model accuracy to $76.7$\% while the model accuracy under benchmark attack schemes remains as high as $87.1$\%.
\end{enumerate}

In summary, we design an effective attack scheme that selects a set of clients for uplink and/or downlink jamming attacks. We show that the downlink attack is more effective than the uplink attack and the combined attack (both downlink and uplink attack) is the best for the adversary in general. Overall, there is no significant need to change the set of clients to attack over time. Moreover, we show how to extend this attack to more general settings where different clients have different processing speeds and attack success probabilities.

The rest of the paper is organized as follows.
Section~\ref{sec:federated} summarizes the FL process and presents its application for wireless signal classification.
Section~\ref{sec:attack} presents attack schemes on FL by specifying how to select clients when jamming their local model updates to the server or the global model update from the server, or both. Section~\ref{sec:attack-dynamic} consider the case when a dynamic set of clients is attacked over time.  Section~\ref{sec:attack-slow} extends the attack on FL to the scenario that clients have different processing powers and thus update local models at different speeds. Section~\ref{sec:attack-prob} considers the case that jamming attacks may not always cause transmission failures and therefore presents how to design the attack on FL with consideration of attack success probabilities. Section~\ref{sec:conc} concludes this paper.

\section{Federated Learning}
\label{sec:federated}
In this paper, we consider FL for one server and $N$ clients (see Fig.~\ref{fig:FL}). Each client has its own data and trains its own model in each round. Clients send their models to the server over wireless channels. The corresponding over-the-air transmissions of the clients and the server are separated over time or frequency (yielding no interference among each other). The server builds the global model by the federated averaging (FedAvg) algorithm. Then, the server broadcasts this model to all clients in one transmission (again over the air). Clients update their models by this global model. This completes one round for FL. The learning process runs over many rounds to build a global model.
Algorithm~\ref{alg:FL} shows the details of FL with FedAvg. Note that there are many variations on the termination criterion, the local training and the averaging processes. The attack schemes in this paper are designed for a general FL algorithm without any assumption on a particular FL implementation. To measure the performance, we use the following setting:
\begin{enumerate}
	\item FL is run for $R$ rounds.
	\item The local training in each round uses all available local data. \item
	A simple averaging function $w = \frac{1}{N} \sum_{i=1}^N w_i$ is used for FedAvg, where $w_i$ is the weight matrix for the model of client $i$.
\end{enumerate}

The benefits of FL include privacy (clients do not share their individual data with the server) and efficiency (individual client model has less data than all data available to all clients). We consider FL in a wireless network, where the server and each client are represented by one individual (network) node and communicate with each other over wireless channels, namely uplink communications from each client to the server and downlink communications from the server to each client.

\begin{algorithm}
    \caption{FL algorithm.}
    \label{alg:FL}
    \begin{algorithmic}[1]
        \STATE Repeat the following steps until the termination criterion is met.

        \STATE \begin{itemize}
        	\item Each client trains its local model using its own data.
        	\item Each client $i$ sends its local model (weights $w_i$) to the server.
        	\item The server computes a global model by $w = \text{FedAvg}(w_1,w_2, \cdots, w_N) = \frac{1}{N} \sum_{i=1}^N w_i$.
            \item The server sends the global model weights $w$ to all clients and each client $i$ updates its local model by $w_i=w$.
        	 \end{itemize}
    \end{algorithmic}
\end{algorithm}

We apply FL for a wireless signal classification problem. There are background transmissions, where data is transmitted using either QPSK or BPSK. 
There are $N$ spectrum sensors (each corresponding to a client in FL) that collect I\&Q data at different locations. The channel induces path loss depending on distance between transmitter and receiver, and a random phase shift. We assume client locations are randomly distributed. We also consider random noise with fixed power at the receiver. The I\&Q data is analyzed as phase shifts and powers at each client. A group of analyzed sensing results ($16$ phase shifts and $16$ powers, corresponding to $32$ features) is used to build one sample at each client. The label is either $0$ (`BPSK') or $1$ (`QPSK'). The server aims to build a general classifier that can detect QPSK or BPSK signals collected at different locations.

A feedforward neural network (FNN) model is used at each client and the server. The FNN properties are shown in Table~\ref{table:FFN}. The local model (namely, weights) of the FNN is different for each client. 

\begin{table}
	\caption{FNN properties.}
	\centering
	{\small
		\begin{tabular}{c|c}
		Input size & 32 \\ \hline
		Output layer size & 2 \\ \hline
Hidden layer sizes & $128, 64, 32$ \\ \hline
Dropout rate & $0.2$ \\ \hline
Activation function & Relu (hidden layer)\\ & Softmax (output layer) \\ \hline
Loss function & Crossentropy \\ \hline
Optimizer & RMSprop \\ \hline
Number of parameters & 14,626
		\end{tabular}
	}
	\label{table:FFN}
\end{table}

For performance evaluation, we consider $N=10$ spectrum sensors, each with $1000$ samples. Another $1000$ samples including samples from all sensors are used to check the performance of the global model. The accuracy of the global model (trained by FL) after $R=100$ FL rounds is measured as $94.4$\%. The local model accuracy after $100$ FL rounds, the signal-to-noise-ratio (SNR) of received signals, and the channel phase shift of each client are shown in Table~\ref{table:local}, where phase angle is within $[0, 2\pi]$. 
Note that the model accuracy depends on both the received power and the channel phase shift, and high received power does not directly imply high model accuracy as the phase shift may reduce the accuracy.

\begin{table*}
	\caption{The local model accuracy when there is no attack.}
	\centering
	{\small
		\begin{tabular}{c|c|c|c|c|c|c|c|c|c|c}
			Client & 1 & 2 & 3 & 4 & 5 & 6 & 7 & 8 & 9 & 10 \\ \hline \hline
			Accuracy & 89.8\% & 78.4\% & 95.1\% & 92.0\% & 86.8\% & 64.7\% & 89.3\% & 64.3\% & 82.7\% & 92.0\% \\ \hline
            SNR (dB) & 9.34 & 15.75 & 13.04 & 13.87 & 10.27 & 10.40 & 13.22 & 9.59 & 12.32 & 8.70 \\ \hline
            Phase shift & 1.28 & 0.43 & 5.18 & 1.93 & 4.37 & 3.09 & 5.42 & 0.77 & 2.44 & 9.40
		\end{tabular}
	}
	\label{table:local}
\end{table*}

\section{Jamming Attack on Federated Learning}
\label{sec:attack}

We now consider different jamming attacks on FL. An adversary can jam the transmission from a client to the server (uplink) or from the server to a client (downlink), or both. When the uplink attack is launched, the server can only update the global model based on the received client models (that are not jammed). As a consequence, the global may miss important information from clients under this attack. When the downlink attack is launched, the client under attack does not have the updated global model and thus can further update its model based on its previous local model only. As a consequence, its model does not have new information from the server (or other clients). There is a certain budget allocated for attacks, i.e., the average number of jamming actions per round, and the adversary must operate within this attack budget. The objective of the adversary is to reduce the global model accuracy (at the end of fixed time period) subject to the attack budget.

\begin{table*}
	\caption{The global model performance if one client is attacked.}
	\centering
	{\small
		\begin{tabular}{c|c|c|c|c|c|c|c|c|c|c}
Attacked client & 1 & 2 & 3 & 4 & 5 & 6 & 7 & 8 & 9 & 10 \\ \hline \hline
Uplink attack & 93.0\% & 94.4\% & 93.6\% & 93.6\% & 92.1\% & 85.5\% & 92.9\% & 84.8\% & 93.7\% & 90.2\% \\ \hline
Downlink attack & 94.0\% & 82.4\% & 63.1\% & 68.1\% & 93.8\% & 88.2\% & 89.2\% & 96.2\% & 87.1\% & 95.1\%
		\end{tabular}
	}
	\label{table:oneclient}
\end{table*}

First, we assume that the adversary attacks the same set of clients, once selected, each client has the same processing speed (so that it can send a local update to the server at each FL round), and a transmission always fails if jammed. We will relax these assumptions in subsequent sections.
Suppose the attack budget dictates that the adversary can attack $M$ clients per round on average. To determine these $M$ clients, we first analyze the case that the adversary always attacks one client and sort them in a list based on the global model accuracy (from worst to best) under the attack. Then, we choose the top $M$ clients to attack.

Table~\ref{table:oneclient} shows the results for the uplink attack (namely, the adversary jams the transmission of a particular client to the server) and downlink attack (namely, the adversary jams the transmission of the server to a particular client). The smallest model accuracy of $84.8$\% is achieved by attacking client $8$ for the uplink attack. On the other hand, the smallest accuracy $63.1$\% is achieved by attacking client $3$ for the downlink attack. Moreover, Table~\ref{table:local} shows that when there is no attack, client $8$ has the worse local model and client $3$ has the best local model after $R=100$ rounds.
Thus, the sets of clients for uplink and downlink attacks should be different and these sets can be determined by the local model accuracies. For uplink updates, the server needs to have a diverse set of data sources from clients. Thus, it is best to attack a client with low local model accuracy since it probably has some unique data not represented by the other clients' data.
For downlink updates, it is necessary to achieve fast convergence. Thus, it is best to attack a client with high local model accuracy to prevent fast convergence (if downlink updates are available).

We further perform correlation analysis on two lists of clients sorted by their local model accuracy (from worst to best) and by the global model accuracy (from worst to best) if these clients are under attack. We find that these two rankings are almost the same if the attack is an uplink attack while these two rankings are almost in the reverse order if the attack is a downlink attack.
Note that the global model accuracy when the adversary attacks $k$ clients is determined by the set of first $k$ elements in a list. Thus, we compare two lists by considering the difference between two sets $\{ x_1, x_2, \cdots, x_k\}$ and $\{ y_1, y_2, \cdots, y_k\}$ for $k = 1, 2, \cdots, N-1$, where the difference is the number of elements included in one set but not included in the other set and its value is in $[0,k]$. We find that the difference is up to $2$ for local accuracy (from worst to best) list and the uplink list, and $3$ for local accuracy (from best to worst) list and the downlink list. This small difference indicates that the local accuracy can be used to select the clients for attack.
Based on this observation, we build the attack scheme in Algorithm~\ref{alg:ideal} when the adversary attacks the same set of $M$ clients per round.

\begin{algorithm}
    \caption{The algorithm for the idealized jamming attack on FL.}
    \label{alg:ideal}
    \begin{algorithmic}[1]
        \STATE Run FL for $R$ rounds without attack and obtain the local model accuracy for each client.

        \STATE Start the FL process from round $1$ again. The adversary performs the attack on selected clients from round $1$.

        \STATE \begin{itemize}
        	\item For the uplink attack, the adversary selects $M$ clients with low accuracy.
        	\item For the downlink attack, the adversary selects $M$ clients with high accuracy.
        	\item For the joint uplink and downlink attack, the adversary uses the above two sets of selected clients for uplink and downlink attacks, respectively.
        	 \end{itemize}
    \end{algorithmic}
\end{algorithm}

Note that the attack scheme in Algorithm~\ref{alg:ideal} is not necessarily practical and therefore some design changes are needed to launch a practical attack, as discussed below.

\begin{itemize}
	\item Algorithm~\ref{alg:ideal} is not practical since if FL is already run for $R=100$ rounds, the global model has high accuracy, namely $94.4$\%. Then, it is too late to launch any attack. The adversary cannot make the system rollback to round $1$ and launch its attack from round $1$ on. The adversary can wait for $S$ rounds (where $S>1$ is a small number), collect local models updated at the $S$th round, and then selects clients to launch attacks. Although a small number of rounds cannot yield good local and global models (namely, the ones with high accuracy when used for testing), it provides guidance for the ranking among clients. The ranking converges quickly while each local model's accuracy changes over time. For numerical results, we set $S=3$.
	\item Another practical issue with the attack scheme in Algorithm~\ref{alg:ideal} is that the adversary does not have any client's local data and thus cannot check the client model accuracy. One approach to circumvent this issue is that the adversary compares the difference between the client model and the global model (at the server) and assumes that a client model with large difference has low accuracy.
\end{itemize}

The practical issues and the necessary design changes described above lead to the attack scheme presented in Algorithm~\ref{alg:attack}.

\begin{algorithm}
    \caption{The algorithm for the practical jamming attack on FL.}
    \label{alg:attack}
    \begin{algorithmic}[1]
        \STATE The adversary waits for $S$ rounds and collects client models by overhearing the transmissions.

        \STATE The adversary calculates the global model based on client models and then calculates the difference between the global model and each client model.

        \STATE \begin{itemize} \item For the uplink attack, the adversary selects $M$ clients with large difference. \item For the downlink attack, the adversary selects $M$ clients with small difference. \item  For the joint uplink and downlink attack, the adversary uses the above two sets of selected clients for uplink and downlink attacks, respectively. \end{itemize}

        \STATE The adversary performs the attack on selected clients from round $S+1$.
    \end{algorithmic}
\end{algorithm}

\begin{figure}
  \centering
  \includegraphics[width=0.925\columnwidth]{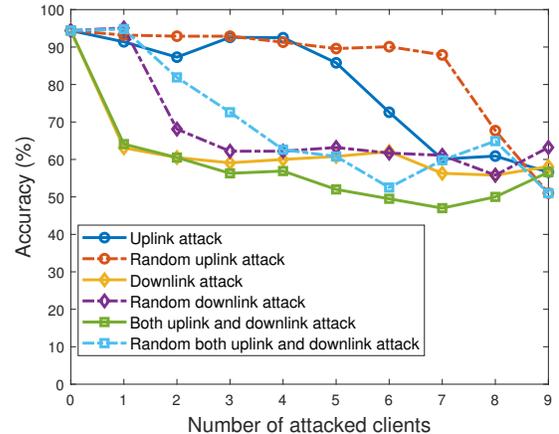}
  \caption{FL attack performance.}
  \label{fig:attack}
\end{figure}

In addition to trying to overhear the local model directly, an adversary can also learn the local model's accuracy indirectly. Suppose that clients use their local models to make spectrum access decisions, e.g., QPSK may be used by primary users while BPSK is used by secondary users. Then, a client may compete for spectrum access if BPSK is detected (and vacate the channel if QPSK is detected). This way, the adversary can observe client actions to learn the local model accuracy.

Note that no matter the adversary uses the difference between two models or the client actions to estimate the model accuracies, it may not know the exact model accuracies. As the adversary needs the ranking of model accuracies, not exact values, it may end up with a good ranking for launching the attack, as we observe in numerical results discussed below.

Figure~\ref{fig:attack} shows the global model accuracy under (i) uplink attack, (ii) downlink attack, and (iii) the joint uplink and downlink attack (when both uplink and downlink attacks are launched). Attacking both the uplink and the downlink is most effective but the level of jamming efforts, namely the number of times the (uplink or downlink) transmissions are jammed, is increased comparing with only attacking the uplink or the downlink (which is taken care of by imposing the attack budget). Attacking the downlink is better than attacking the uplink as the loss of global model update is more critical than the loss of local model updates. Note that the accuracy may not always decrease when the adversary attacks more clients because the incremental contribution of some clients on the global model may not be positive. The designed attack scheme is compared with a random attack scheme, where the adversary randomly selects which clients to attack. Our results show that this random attack scheme cannot effectively reduce the global model accuracy.

\begin{table}
	\caption{Performance of attacking different clients in each round.}
	\centering
	{\small
		\begin{tabular}{c|c|c|c|c|c}
$(M,K)$ & (1,1) & (1,2) & (2,2) & (2,3) & (2,4) \\ \hline \hline
Uplink & 91.4\% & 90.3\% & 87.3\% & 94.0\% & 93.6\% \\ \hline
Downlink & 63.1\% & 93.5\% & 60.5\% & 91.4\% & 93.9\% \\ \hline
Both & 64.1\% & 92.1\% & 60.5\% & 89.3\% & 91.9\%
		\end{tabular}
	}
	\label{table:dynamic}
\end{table}

\section{Jamming Attack on Different Clients in Each FL Round}
\label{sec:attack-dynamic}
So far, we limit that the adversary attacks the same set of clients (a fixed set of clients, once selected) in all FL rounds. The question is whether the adversary can achieve better attack performance by attacking different clients in each round. We consider the case that the adversary can attack up to $K$ clients in a round (these clients can be selected based on the local model accuracy). The average number of attacked clients is still $M$ so that we can have a fair comparison with previous results. We denote this scheme as $(M,K)$ and previous scheme (with fixed set of attacked clients) becomes a special case denoted as $(M,M)$. Table~\ref{table:dynamic} shows the performance of such attack, where we compare with previous schemes with the same $M$ value. Except the uplink attack for schemes $(1,1)$ and $(1,2)$, the attack scheme with $(M,K)$ is worse than the one with $(M,M)$. Hence, it is better in most of the cases to attack the same set of clients in all rounds. Overall, if the adversary does not attack a client in all rounds, there is an opportunity for a model update transmitted to or from that client in some rounds, which improves the global model accuracy over time.

\begin{table}
	\caption{Performance comparison for continuously ranking clients for the attack on FL.}
	\centering
	{\small
		\begin{tabular}{c|c|c|c|c}
Attack & \multicolumn{2}{c}{Sensing once} & \multicolumn{2}{|c}{Sensing at each round} \\ \cline{2-5}
type & $M=1$ & $M=2$ & $M=1$ & $M=2$ \\ \hline \hline
Uplink & 91.4\% & 87.3\% & 93.1\% & 92.0\% \\ \hline
Downlink & 63.1\% & 60.5\% & 94.1\% & 93.6\% \\ \hline
Both & 64.1\% & 60.5\% & 89.9\% & 88.4\%
		\end{tabular}
	}
	\label{table:dynamic2}
\end{table}

The attack scheme in Algorithm~\ref{alg:attack} uses the local model at round $S$ to select clients. One question is whether the adversary should keep checking the client model performance to select potentially different clients for the attack. Table~\ref{table:dynamic2} shows the results of this scheme (in the fourth and fifth column). The new scheme cannot achieve good performance. The reason is that except the first few sensing results, the subsequent sensing results are obtained under the attack and cannot provide reliable information to select which clients to attack.

\section{Jamming Attack on Clients with Different Processing Speeds}
\label{sec:attack-slow}

Suppose that clients may have different processing powers that are translated to different processing speeds. As a result, a client with high processing power can update its local model and send it to the server in every FL round, while a client with low processing power needs multiple rounds to update its local model and then sends it to the server once for every multiple FL rounds. We denote $s_i$ as the number of FL rounds that client $i$ takes to update its local model.

The adversary waits for $S \times \max_i s_i$ rounds (we set $S=3$ for numerical results) and then collects client local models. It again selects clients to attack based on local model accuracy, which is the same as the attack scheme in Algorithm~\ref{alg:attack}. The difference is that, once the adversary identifies slow clients, there is no need to jam them in every round. As a consequence, the attack budget (averaged over time) to jam a slow client is $\frac{1}{s_i}$. That is, the total budget to jam a set $C$ of clients is $\sum_{i=1}^N \frac{1}{s_i}$. Thus, we end up with the attack scheme presented in Algorithm~\ref{alg:slow} for the jamming attack on clients with different processing speeds.

\begin{algorithm}
    \caption{The algorithm for the jamming attack on FL when clients have different processing speeds.}
    \label{alg:slow}
    \begin{algorithmic}[1]
        \STATE Suppose each client $i$ can update its local model for every $s_i$ rounds.

        \STATE Wait for $S \times \max_i s_i$ rounds and collect client models by overhearing the transmissions.

        \STATE Calculate the global model based on client models and calculate the difference between the global model and each client model.

        \STATE Denote the set of selected clients as $C$. The attack budget is $\sum_{i=1}^N \frac{1}{s_i} \le M$.
        \begin{itemize} \item For the uplink attack, the adversary selects clients with large difference. \item For the downlink attack, the adversary selects clients with small difference. \item For the joint uplink and downlink attack, the adversary uses the above two sets of selected clients for uplink and downlink attacks, respectively. \end{itemize}

        \STATE The adversary performs the attack on selected clients from round $S \times \max_i s_i  + 1$.
    \end{algorithmic}
\end{algorithm}

We compare this attack scheme with the following benchmark attack schemes:
\begin{enumerate} [start=1,label={A\arabic*.}]
	\item The adversary does not know the existence of slow clients and attacks all selected clients in every round.
	\item The adversary first selects fast clients and then slow clients to attack.
	\item The adversary first selects slow clients and then fast clients to attack.
	\item The adversary randomly selects clients to attack.
\end{enumerate}

\begin{figure}
  \centering
  \includegraphics[width=0.925\columnwidth]{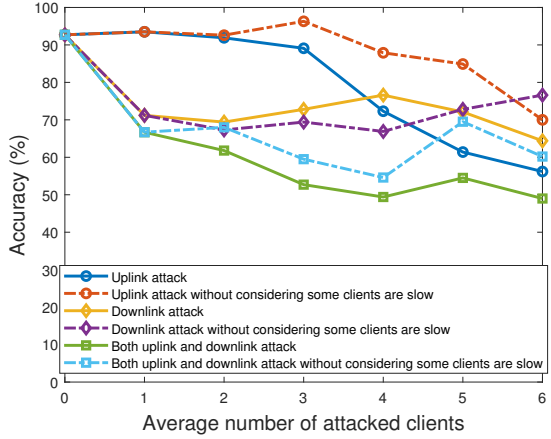}
  \caption{The comparison of FL attack performance  with benchmark attack scheme A1.}
  \label{fig:fastslow1}
\end{figure}

\begin{figure}
  \centering
  \includegraphics[width=0.925\columnwidth]{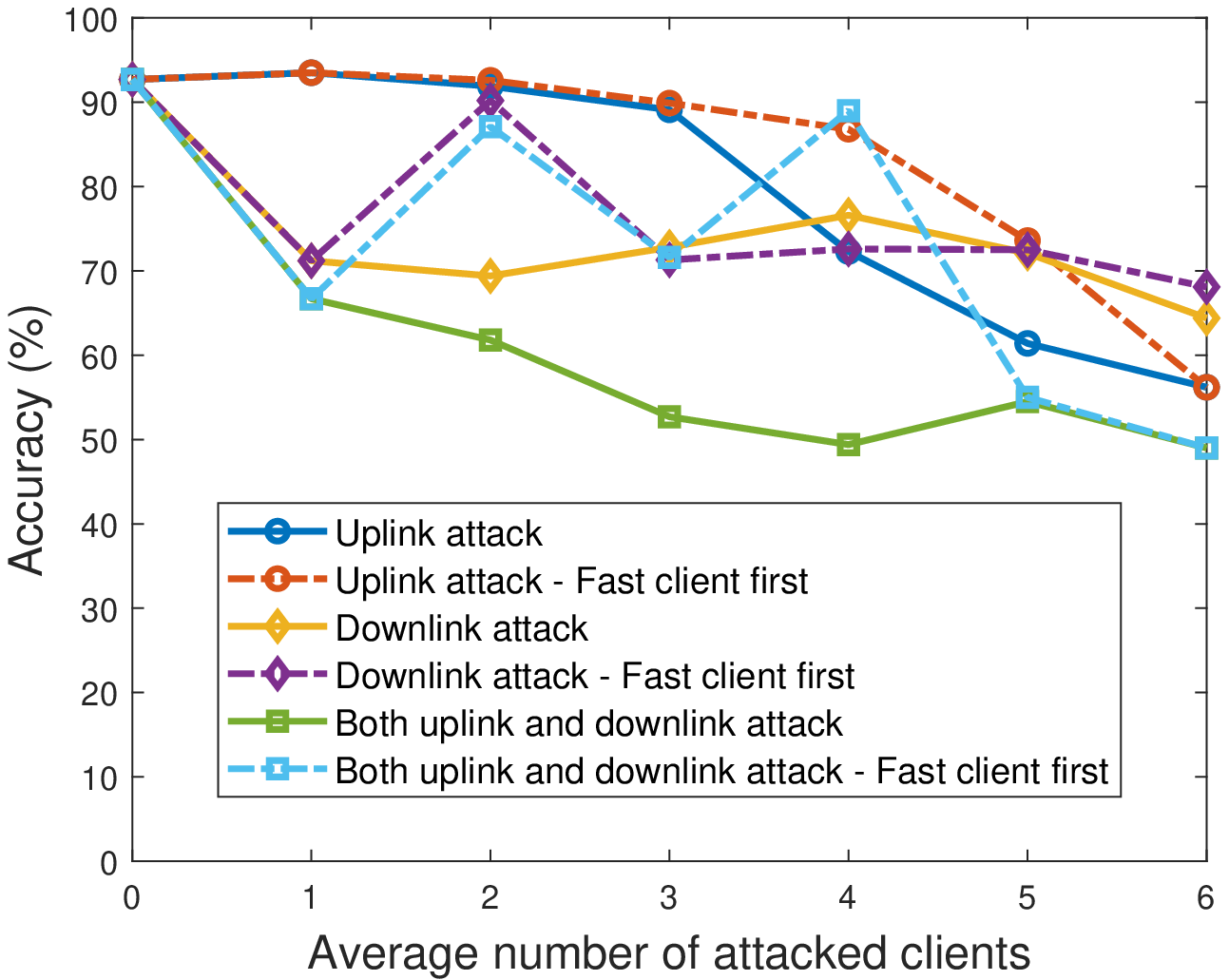}
  \caption{The comparison of FL attack performance  with benchmark attack scheme A2.}
  \label{fig:fastslow2}
\end{figure}

\begin{figure}
  \centering
  \includegraphics[width=0.925\columnwidth]{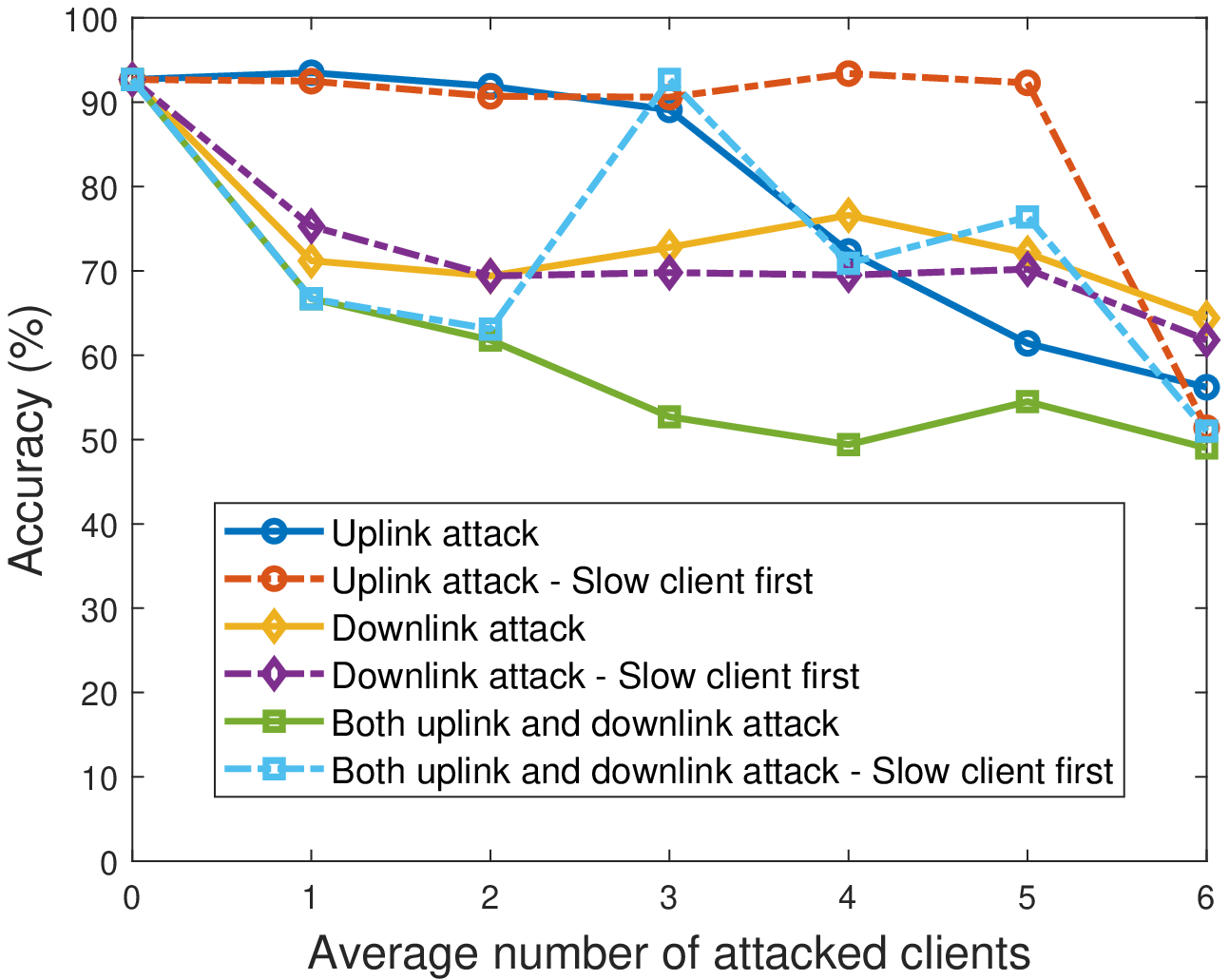}
  \caption{The comparison of FL attack performance  with benchmark attack scheme A3.}
  \label{fig:fastslow3}
\end{figure}

\begin{figure}
  \centering
  \includegraphics[width=0.925\columnwidth]{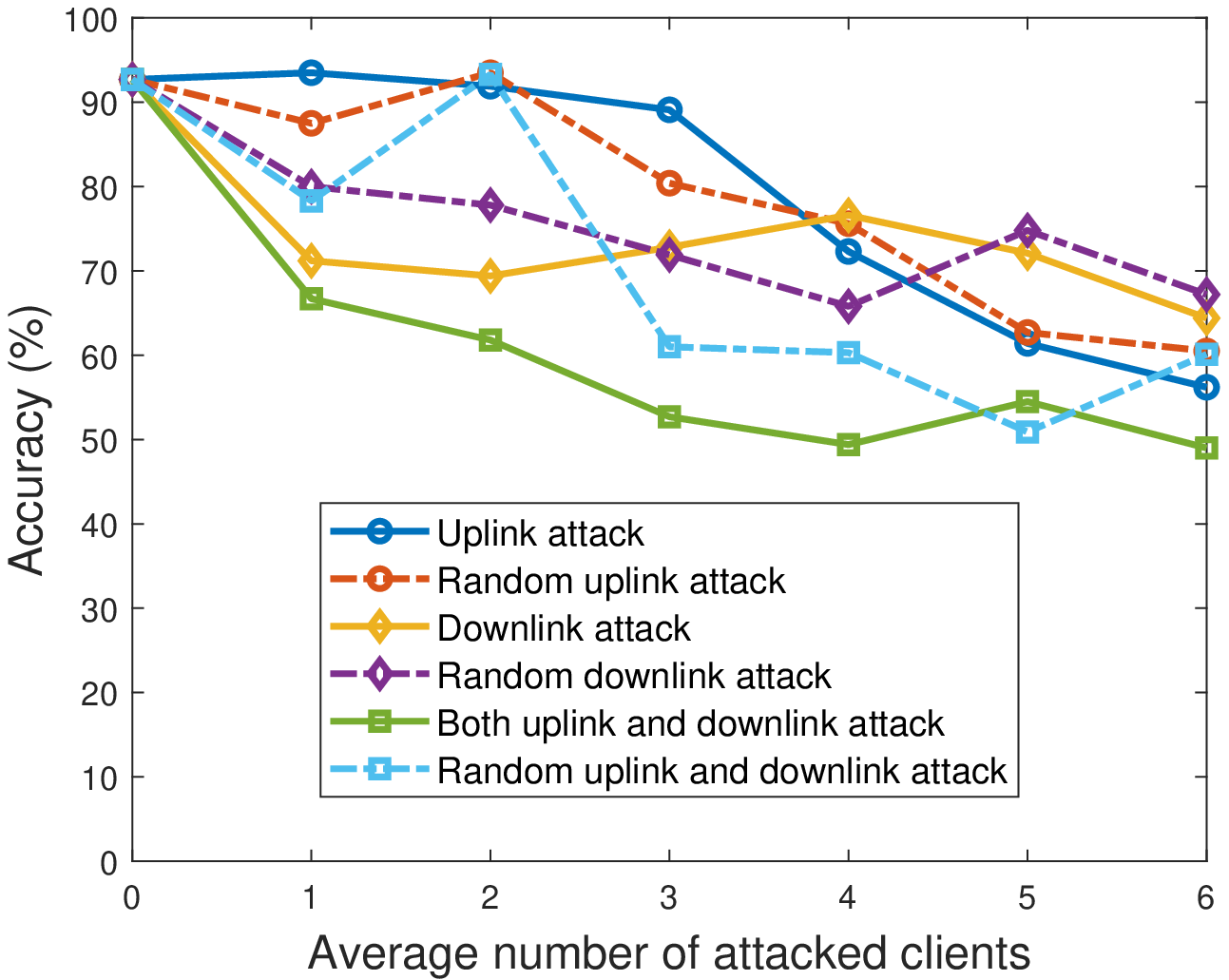}
  \caption{The comparison of FL attack performance  with benchmark attack scheme A4.}
  \label{fig:fastslow4}
\end{figure}

We consider the case that clients $1$ to $6$ are slow clients with $s_i= 2$ for $i=1,\cdots,6$ and clients $7$ to $10$ are fast clients with $s_i=1$ for $i = 7, \cdots, 10$. When there is no attack, the global model has accuracy $92.7$\% after $R=100$ rounds.
Figures~\ref{fig:fastslow1} - \ref{fig:fastslow4} show the comparison of the attack scheme in Algorithm \ref{alg:slow} with each benchmark scheme when we change the average number of attacked clients per round from $0$ to $6$. Figure~\ref{fig:fastslow1} shows that the attack scheme in Algorithm~\ref{alg:slow} is better than benchmark attack scheme A1 for most of cases, since this benchmark scheme does not know the existence of slow clients and thus cannot fully utilize the budget to attack. Figures~\ref{fig:fastslow2} and \ref{fig:fastslow3} show that the attack scheme in Algorithm~\ref{alg:slow} is better than benchmark attack schemes A2 and A3 for most of cases, since these benchmark attack schemes limit the selection sequence among clients.
Figure~\ref{fig:fastslow4} shows that the attack scheme in Algorithm~\ref{alg:slow} is better than benchmark attack scheme A4 for most of cases, since this benchmark attack scheme randomly selects clients to attack and does not account for model accuracies.

\begin{table*}
	\caption{The attack success probabilities of clients.}
	\centering
	{\small
		\begin{tabular}{c|c|c|c|c|c|c|c|c|c|c}
Client & 1 & 2 & 3 & 4 & 5 & 6 & 7 & 8 & 9 & 10 \\ \hline \hline
Probability & 82.69\% & 96.95\% & 95.28\% & 85.10\% & 89.91\% & 88.99\% & 93.03\% & 95.77\% & 81.88\% & 80.57\%
		\end{tabular}
	}
	\label{table:prob}
\end{table*}

\section{Jamming Attack on Clients with Different Attack Success Probabilities}
\label{sec:attack-prob}

So far, we assumed that the jamming attack is always successful in terms of preventing the update from reaching the client and the server in the downlink and uplink attack, respectively. The next scenario that we consider in this section is that when the adversary attacks a client or a server, the underlying jamming action may not be always successful, e.g., the adversary can jam a physically close client or server with a high probability due to higher interference (jamming) power experienced at the client or server. The transmission of a local or global model update fails if the signal-to-noise-and-interference ratio (SINR) experienced (by a server or a client) is less than a threshold. Note that the SINR is a random variable and depends on the distributions of random wireless channels between the server and each client, as well as the channels from the jammer to the server and each client. We assume locations of the clients and the jammer are randomly distributed. Therefore, we assume a probabilistic update reception model based on the SINRs under the jamming attack. In the meantime, we assume that the processor speed is the same again for all clients so that the attack success probability and model accuracy are the two factors that we consider when designing the jamming attack on FL.
	
For the uplink attack, Table~\ref{table:prob} shows the success probability for each client (depending on channel effects) if it is under attack.
For this case, the attack scheme that considers the model accuracy only may not be able to achieve the best attack performance. Another attack scheme that considers the attack success probability only may not be able to achieve the best performance either as it would ignore the model characteristics of clients. Therefore, it is essential to combine the model accuracy and the attack success probability when the adversary determines how to select which clients to attack.
For example, the best client for the downlink attack is client $3$ based on model accuracy while based on attack success probability, the best client to attack is client $1$ (the one with the largest attack success probability).
Thus, we need to consider the model accuracy and the attack success probability jointly to select clients to attack.

For the uplink attack, the adversary aims to select clients such that the diversity of remaining clients is minimized. Denote the difference between client $i$'s model and the global model as $d_i$ and the attack success probability as $p_i$. The diversity (average difference) is $\frac{\sum_j d_j}{N}$ if the attack fails (with probability $1-p_i$) and $\frac{\sum_{j \ne i} d_j}{N-1}$ if the attack succeeds (with probability $p_i$). Minimizing this diversity is equivalent to maximizing $p_i \left(d_i- \frac{\sum_j d_j}{N} \right)$.
For the downlink attack, the adversary selects clients with small $\frac{d_i}{p_i}$, where $d_i$ denotes the difference between client $i$'s model and the global model and $p_i$ denotes the attack success probability of client $i$. Then, we end up with the attack scheme presented in Algorithm~\ref{alg:prob}. Note that if $p_i = 1$ for each client $i$, Algorithm~\ref{alg:prob} reduces to Algorithm~\ref{alg:attack}.

\begin{algorithm}
    \caption{The algorithm for the jamming attack on FL with consideration of attack success probability.}
    \label{alg:prob}
    \begin{algorithmic}[1]
        \STATE Wait for $S$ rounds and collect client models by overhearing.

        \STATE Calculate the global model based on client models and calculate the difference between the global model and each client model.

        \STATE The adversary can select $M$ clients for attacks.
        \begin{itemize}
        \item For the uplink attack, the adversary selects clients with large $p_i (d_i- \frac{\sum_j d_j}{N})$.
            
        \item For the downlink attack, the adversary selects clients with small $\frac{d_i}{p_i}$.
        \item For the joint uplink and downlink attack, the adversary uses the above two sets of selected clients for uplink and downlink attacks, respectively.
    	\end{itemize}

        \STATE The adversary performs the attack on selected clients from round $S+1$.
    \end{algorithmic}
\end{algorithm}

\begin{figure}
  \centering
  \includegraphics[width=0.925\columnwidth]{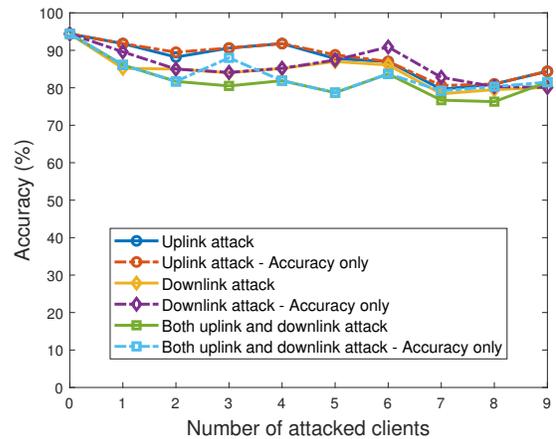}
  \caption{The comparison of FL attack performance  with benchmark attack scheme A5.}
  \label{fig:prob1}
\end{figure}

\begin{figure}
  \centering
  \includegraphics[width=0.925\columnwidth]{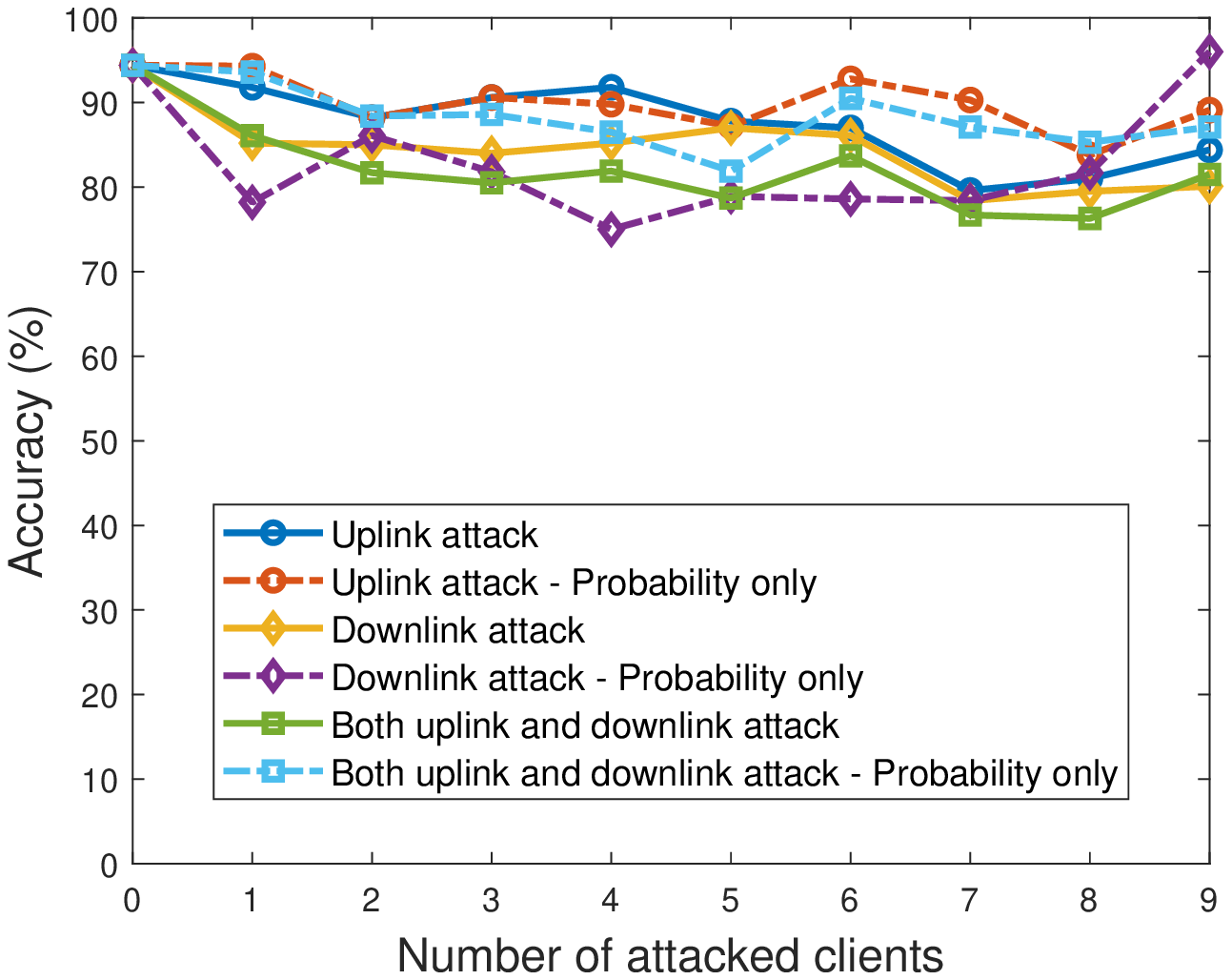}
  \caption{The comparison of FL attack performance  with benchmark attack scheme A6.}
  \label{fig:prob2}
\end{figure}

\begin{figure}
  \centering
  \includegraphics[width=0.925\columnwidth]{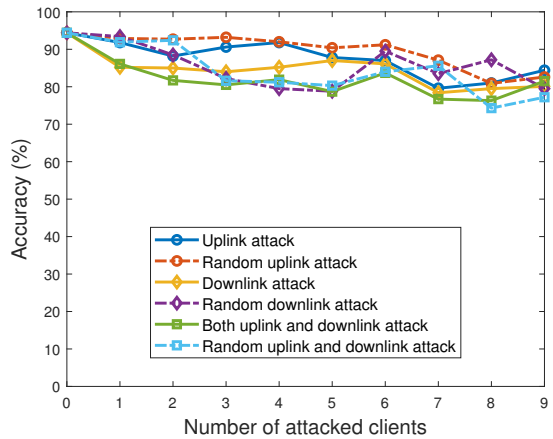}
  \caption{The comparison of FL attack performance  with benchmark attack scheme A7.}
  \label{fig:prob3}
\end{figure}

We compare this attack scheme with the following  benchmark attack schemes:
\begin{enumerate} [start=5,label={A\arabic*.}]
	\item The adversary considers accuracy only to select clients.
	\item The adversary selects clients with high attack success probability.
	\item The adversary randomly selects clients to attack.
\end{enumerate}
Figures~\ref{fig:prob1}--\ref{fig:prob3} show the comparison of the designed scheme with each benchmark scheme.
Figure~\ref{fig:prob1} shows that the attack scheme in Algorithm~\ref{alg:prob} is better than benchmark attack scheme A5 for most of cases, since this benchmark attack scheme only considers model accuracy.
Figure~\ref{fig:prob2} shows that the attack scheme in Algorithm~\ref{alg:prob} is better than benchmark attack scheme A6 for most of cases (except the downlink attack), since this benchmark attack scheme only considers attack success probability.
Figure~\ref{fig:prob3} shows that the attack scheme in Algorithm~\ref{alg:prob} is better than benchmark attack scheme A7 for most of cases, since this benchmark attack scheme randomly selects clients to attack and does not account for model accuracy or attack success probability.

\section{Conclusion}
\label{sec:conc}

In this paper, we applied FL for a wireless signal classification task and studied various jamming attacks on FL that is performed over a wireless network (where the client and the server communicate over wireless channels). We first showed that without an attack, the global model trained by FL can classify signals with different modulations (received at different locations with different channel gains and phase shifts) with high accuracy. We considered an adversary to launch jamming attacks on FL with the objective of reducing the classification accuracy. In particular, the attack can either jam the local model updates to the server (uplink attack) or jam the global model updates from the server (downlink attack), or jam both. In all cases, we imposed an attack budget in terms of the average number of attacked clients and designed schemes to select clients for uplink/downlink attacks based on the client local model accuracy when there is no attack. To make this attack scheme practical, the adversary uses the overheard local models or the observed client actions based on local models after several rounds to predict the ranking of local model accuracies. We showed that this attack  is very effective and reduces the classification accuracy significantly. We further studied the case that the set of selected clients changes over time and found that this dynamic setting cannot improve the attack performance. Then, we extended the jamming attack to more general settings, i.e., clients have different processing speeds or there are different attack success probabilities for clients, and showed that it can effectively reduce the global model accuracy compared to benchmark attack schemes.

\end{document}